\documentclass[preprint,12pt, a4paper]{elsarticle}

\usepackage{amssymb}
\usepackage{hyperref}
\usepackage{xcolor}
\usepackage{orcidlink}
\usepackage{hypcap}
\usepackage{float}

\usepackage{xcolor}
\usepackage{listings}

\definecolor{codebg}{RGB}{245, 245, 245}      
\definecolor{codecomment}{RGB}{70, 150, 70}   
\definecolor{codekeyword}{RGB}{200, 80, 40}   
\definecolor{codestring}{RGB}{40, 100, 170}   
\definecolor{codefunction}{RGB}{30, 120, 190} 
\definecolor{codetext}{RGB}{30, 30, 30}       

\lstdefinestyle{pythonstyle}{
    language=Python,
    backgroundcolor=\color{codebg},
    basicstyle=\ttfamily\scriptsize\color{codetext},
    keywordstyle=\color{codekeyword}\bfseries,
    commentstyle=\color{codecomment},
    stringstyle=\color{codestring},
    identifierstyle=\color{codetext},
    showstringspaces=false,
    breaklines=true,
    frame=single,
    rulecolor=\color{codebg},
    columns=fullflexible,
    keepspaces=true,
    tabsize=4,
    aboveskip=3pt,
    belowskip=3pt
}

\hypersetup{
    colorlinks=true,
    linkcolor=blue,
    citecolor=blue,
    urlcolor=blue,
    breaklinks=true
}

\journal{SoftwareX}

\begin{document}
\renewcommand{\labelenumii}{\arabic{enumi}.\arabic{enumii}}

\begin{frontmatter}



\title{FrontVeg V2: A Training-Free Software Framework for Foreground-Aware Zero-Shot Plant Trait Segmentation in High-Resolution Images of Trellised Crops}



\author[irhs]{Abdoul Djalil Ousseini Hamza\orcidlink{0009-0008-6172-4439}}
\author[laris]{Herearii Metuarea\orcidlink{0009-0008-2716-0617}}
\author[irhs]{Corentin Lothodé\orcidlink{0000-0002-8209-317X}}
\author[gafl]{Morgane Roth\orcidlink{0000-0002-5244-4215}}
\author[svqv]{Jacem Ben Hamden \orcidlink{0009-0003-7910-9737}}
\author[svqv]{Eric Duchêne\orcidlink{0000-0003-2712-1892}}
\author[ueav]{Lionel Ley}
\author[uea]{David Alletru\orcidlink{0009-0006-6238-6123}}
\author[laris]{David Rousseau\orcidlink{0000-0002-7935-1609}
    \corref{cor1}}

\cortext[cor1]{Corresponding author.}
\ead{david.rousseau@univ-angers.fr}

\address[irhs]{IRHS, INRAE, Univ. Angers, Angers, France}
\address[laris]{LARIS, Univ. Angers, Angers, France}
\address[gafl]{GAFL, INRAE, Montfavet, France}
\address[svqv]{SVQV, UMR1131, INRAE, Univ. Strasbourg, Colmar, France}
\address[ueav]{UEAV, INRAE, Colmar, France}
\address[uea]{UEA, UMR0393, INRAE, Toulenne, France}

\begin{abstract}
FrontVeg V2 is an open-source, training-free software framework for foreground-aware zero-shot segmentation of plant traits in high-resolution images of trellised crops. The pipeline combines monocular depth estimation, automatic foreground extraction using Valley-Aware Depth Thresholding, tiled zero-shot segmentation, Graph-Based Mask Assembly, and geometry-aware fusion. This design enables plant organs and disease symptoms to be segmented while reducing detections arising from neighboring vegetation rows. The current implementation integrates Depth Anything V2 (DAV2) and SAM3 and can be used through both command-line batch processing and a Napari graphical interface. FrontVeg V2 provides a reusable framework for multi-crop, multi-trait digital phenotyping without task-specific model retraining.

\end{abstract}


\end{frontmatter}


\section*{Required Metadata}
All metadata are provided in Table~\ref{tab:metadata}, and the corresponding GitHub repository is publicly available.
\label{}



    \begin{table}[!h]
    \begin{tabular}{|l|p{7cm}|p{7cm}|}
    \hline
    \textbf{Nr.} & \textbf{Code metadata description} & \textbf{Please fill in this column} \\
    \hline
    
    C1 & Current code version & v2.2.0 \\
    \hline
    C2 & Permanent GitHub link to code/repository used for this code version &
    \url{https://github.com/djaliloh/FrontVegV2} \\
    \hline
    C3 & Legal Code License & MIT License \\
    \hline
    C4 & Code versioning system used & Git \\
    \hline
    C5 & Software code languages, tools, and services used &
    Python, PyTorch, Napari \\
    \hline
    C6 & Compilation requirements, operating environments \& dependencies &
    Miniforge/Anaconda, Python 3.10, CUDA 12.1, PyTorch, SAM3,
    Depth Anything V2, napari, NumPy, OpenCV, SciPy, scikit-image,
    scikit-learn, Pillow, pandas, matplotlib \\
    \hline
    C7 & If available Link to developer documentation/manual &
    \url{https://github.com/djaliloh/FrontVegV2} \\
    \hline
    C8 & Support email for questions &
    \url{david.rousseau@univ-angers.fr} \\
    \hline
    
    \end{tabular}
    \caption{Code metadata}
    \label{tab:metadata} 
    \end{table}





\section{Motivation and significance}
Accurate plant trait segmentation is an important step in digital phenotyping, as it provides the basis for measuring traits such as leaf area, fruit number, and disease severity from images. 


In trellised cropping systems, such as vineyards and modern orchards, this task remains challenging because a single side-view image can contain several vegetation rows with very similar visual appearances. As a result, segmentation models can detect organs from neighboring rows together with the target vegetation. 

This is particularly problematic when the objective is to characterize only the foreground crop row. At the same time, phenotyping images are often acquired at high resolution to preserve small structures such as leaves, fruits, and disease symptoms, making direct full-image inference difficult for many vision models. 

Monocular depth foundation models offer a complementary source of information by estimating relative scene depth directly from a single RGB image, without requiring additional depth sensors.
Recent vision foundation models provide strong zero-shot and open-vocabulary segmentation capabilities without task-specific training. Models such as SAM and SAM2 \cite{kirillov2023segment,ravi2024sam2segmentimages}, as well as open-vocabulary approaches such as GroundingDINO+SAM \cite{liu2024grounding,liu2023groundingdino,kirillov2023segment} and SAN \cite{xu2023san}, have reduced the need for task-specific segmentation models. In parallel, monocular depth estimation has advanced through models such as MiDaS, ZoeDepth, Metric3D, Depth Anything, and UniDepth \cite{ranftl2020towards,bhat2023zoedepth,yin2023metric3d,depth_anything_v1,depth_anything_v2,piccinelli2024unidepth}. In agricultural imaging, existing datasets and methods have addressed fruit, leaf, grape cluster, and disease segmentation, including MinneApple \cite{hani2019minneapple}, WGISD \cite{santos2019wgisd}, GrapeSAM \cite{torreslomas2024grapesam}, ViViD-5K \cite{tong2026vivid5k}, PlantSeg \cite{wei2024plantseg}, and disease-specific approaches \cite{rossi2022ldd,kerkech2019vine,kerkech2020vddnet,rous2023apple}. However, these approaches generally focus on a specific crop, trait, disease, sensor, or segmentation task, and do not explicitly use scene geometry to identify the foreground vegetation row. FrontVeg V2 builds on these complementary research directions by combining monocular depth estimation with tiled zero-shot segmentation in a training-free pipeline. The estimated depth is used as a geometric prior to restrict the segmentation to the foreground crop row, while open-vocabulary prompts allow the same framework to target different plant traits and disease symptoms.

FrontVeg V2 is a training-free software pipeline designed for foreground-aware zero-shot segmentation of plant traits in high-resolution images of trellised crops. The software combines monocular depth estimation, automatic foreground extraction, tiled zero-shot segmentation, graph-based mask assembly, and geometry-aware mask fusion. Its main objective is not to replace existing foundation models, but to provide a processing framework that makes their zero-shot predictions more suitable for structured agricultural scenes. The current implementation uses Depth Anything V2 \cite{depth_anything_v2} for monocular depth estimation and SAM3 \cite{carion2025sam3segmentconcepts} for zero-shot instance segmentation, while their model paths are provided externally to the software rather than being fixed in the source code.
\newline

From a scientific perspective, FrontVeg V2 provides a reusable framework for studying how monocular scene geometry can be combined with vision foundation models for plant phenotyping. It allows researchers to evaluate zero-shot segmentation on different crops and plant traits without retraining the complete pipeline for each new task. The configurable prompts, tiling parameters, depth-processing parameters, and externally provided model paths also make it possible to test different models and processing configurations within the same workflow. In this way, the software can be used not only as a practical phenotyping tool, but also as an experimental framework for investigating geometry-aware zero-shot segmentation in agricultural environments.
 
The software is also associated with our research presented in the CVPPA 2026 Workshop at ECCV \cite{hamza2026foregroundaware}, where the proposed framework was evaluated for zero-shot plant trait segmentation in trellised crops using side-view monocular RGB images. This provides an initial research application of the software, while the SoftwareX version is intended to make the complete implementation more accessible and reusable for future agricultural imaging studies. 
\newline


The software was evaluated on RGB images of grapevine, apricot, and apple trees acquired under heterogeneous field conditions. The evaluation dataset comprises 158 images from six geographic locations in France, Spain, Italy, Belgium, and Switzerland, including grapevine images from Colmar and apricot images from Amarine (France), as well as apple images from the DeepPhenoTree--CVPPAData Challenge dataset (DOI: \url{https://doi.org/10.57745/NORPF1}) \cite{deepphenotree2026challenge}. Images were acquired with different camera systems, illumination conditions, and viewpoints. The benchmark covers leaf and fruit segmentation for all three crops, and black rot and downy mildew segmentation for grapevine. Ground-truth masks (Table~\ref{tab:dataset_distribution}) were manually annotated at pixel level using LabelBox \cite{labelbox}, following guidelines established with plant scientists and agronomists. Approximately 200 hours of annotation produced around 10,000 manually delineated object instances.

    \setlength{\tabcolsep}{4pt}
    \begin{table}[!ht]
    \centering
    \caption{Distribution of images and annotated traits across species.}
    \label{tab:dataset_distribution}
    \begin{tabular}{lccccc}
    \hline
    Species &
    Leaf &
    Fruit &
    Black Rot &
    Downy Mildew &
    Total Images \\
    \hline
    Grapevine & 15 & 25 & 25 & 23 & 88 \\
    Apricot Tree & 10 & 25 & -- & -- & 35 \\
    Apple Tree & 10 & 25 & -- & -- & 35 \\
    \hline
    Total & 35 & 75 & 25 & 23 & 158 \\
    \hline
    \end{tabular}
    \end{table}

The proposed framework was evaluated on four phenotypic traits representative of vineyard and orchard monitoring: \emph{leaf} and \emph{fruit} for all crop species, and the two major grapevine diseases, \emph{black rot} and \emph{downy mildew}. These classes correspond to common phenotyping tasks related to canopy characterization, yield estimation, and disease assessment.
\newline

FrontVeg V2 can be used in two main ways. For large-scale experiments, the command-line interface supports directory-based processing and configurable parameters such as the input and output directories, textual prompt, tile size, tile overlap, and depth-processing parameters. This allows the same pipeline to be applied to a collection of images without manual intervention. For interactive analysis, the software is also provided as a Napari plugin \url{https://napari-hub.org/plugins/frontvegv2.html}, allowing users to load an image, configure the main processing parameters, run the segmentation pipeline, and inspect the resulting masks within the Napari environment. The same software implementation is therefore intended to support both batch processing for experiments and interactive inspection of individual images.
\newline

\section{Software description}

FrontVeg V2 is a modular Python software for foreground-aware zero-shot plant trait segmentation in high-resolution agricultural images. Its core library is decoupled from the graphical interface, enabling both batch processing on HPC clusters and interactive local visualization.

\subsection{Software architecture}
\label{sec:architecture}

\begin{figure}[!ht]
\centering
\includegraphics[width=0.9\textwidth]{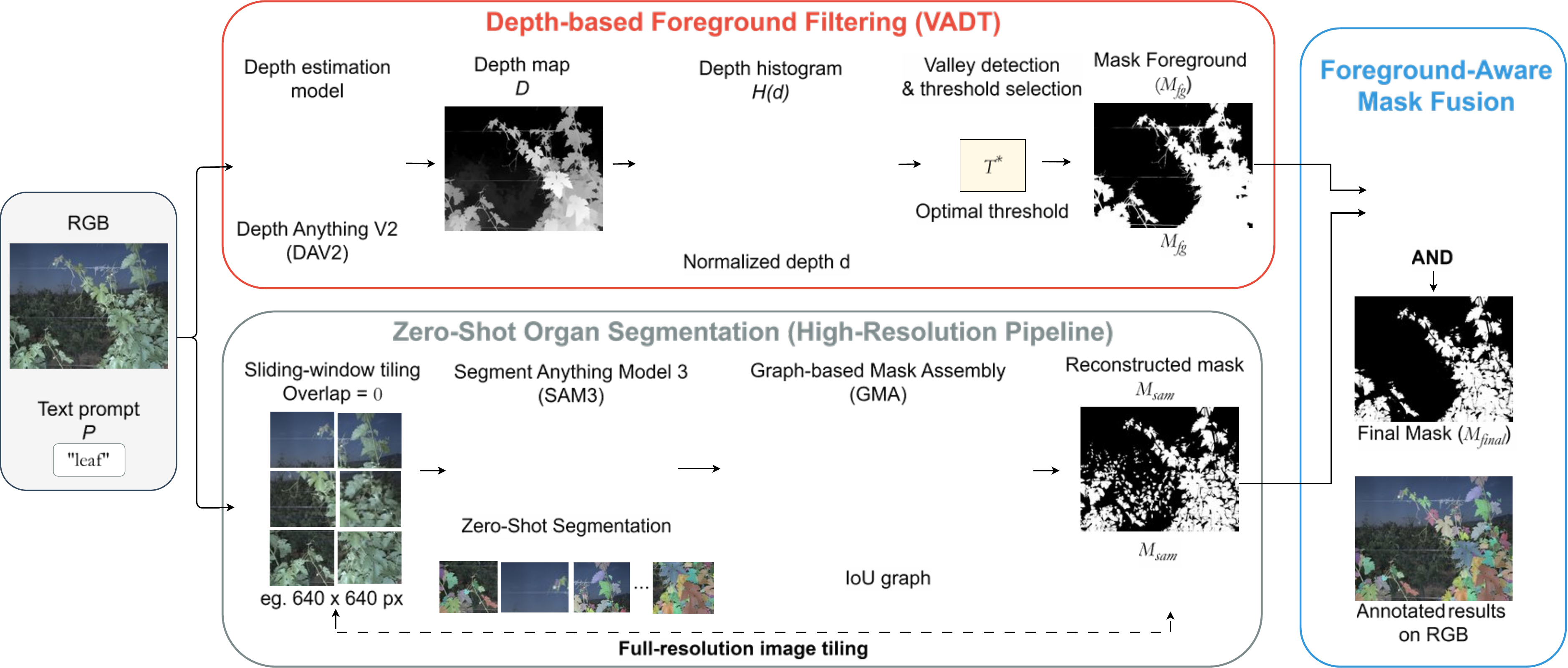}
\caption{Overall architecture of FrontVeg V2. The software processes the input RGB image through a depth-based foreground extraction branch and a tiled zero-shot segmentation branch. The two outputs are combined in the final fusion step to retain plant traits belonging to the foreground vegetation row.}
\label{fig:overallarchi}
\end{figure}

The pipeline is organized into two parallel processing branches followed by a final fusion step. Figure~\ref{fig:overallarchi} illustrates the overall architecture. The software processes a monocular RGB image and a user-defined textual prompt (e.g., \textit{leaf}, \textit{fruit}) to output an instance segmentation map restricted to the foreground vegetation row. The architecture comprises four main modules.

\textbf{Geometry-Aware Foreground Extraction:} 
The first branch extracts a binary foreground mask based on depth information. The software loads a monocular depth foundation model (DAV2 \cite{depth_anything_v2}) from an externally specified path. This design allows users to swap the depth estimator without modifying the pipeline source code. The model estimates a relative depth map from the RGB image.

To automate the selection of the depth threshold, the software applies the Valley-Aware Depth Thresholding (VADT) algorithm. 
The software converts the depth map into a one-dimensional histogram $H(x)$, where $x\in[0,255]$. To suppress local fluctuations, it smooths the histogram with a Gaussian filter, yielding $H_s(x)$. The prominent local maxima form the set $P=\{p_0,p_1,\ldots,p_n\}$. Since the first mode $p_0$ generally corresponds to the distant background, the module discards it. For each consecutive pair of modes $(p_i,p_{i+1})$, the algorithm defines the corresponding valley as:
\[
v_i=\arg\min_{x\in[p_i,p_{i+1}]}H_s(x).
\]
To avoid selecting thresholds from distant regions, the search is limited to the first four candidate valleys $V=\{v_1,v_2,v_3,v_4\}$. The software selects the optimal threshold as:
\[
T_{\mathrm{opt}}=\arg\min_{v\in V}H_s(v).
\]
Thresholding the depth map with $T_{\mathrm{opt}}$ directly yields the binary foreground mask $\mathcal{M}_{fg}$.


\textbf{High-resolution Zero-Shot Segmentation:} 
In parallel, the second branch processes the original RGB image. To preserve small plant structures, the software divides the image into overlapping square tiles. By default, it uses $640 \times 640$ tiles with an overlap factor $O=0.5$. This results in a stride $S$ defined by:
\[
S=L(1-O),
\]
where $L$ is the tile size. Both $L$ and $O$ are configurable by the user. Each tile is independently processed by a vision foundation model (SAM3 in our implementation \cite{carion2025sam3segmentconcepts}) using the user's textual prompt. The software discards fragments smaller than 50 pixels or covering more than 55\% of a tile.

\textbf{Graph-Based Mask Assembly (GMA):} 
Because overlapping tiles generate redundant and fragmented predictions, the software applies a mask assembly module. GMA represents each valid mask fragment as a node in a graph. The module establishes edges between fragments using a two-stage overlap validation. First, it identifies candidate pairs using bounding box Intersection-over-Union ($\mathrm{IoU}_{box}>0.05$). Second, it evaluates these pairs using the mask IoU:
\[
\mathrm{IoU}_{mask}(m_i,m_j)= \frac{|m_i\cap m_j|}{|m_i|+|m_j|-|m_i\cap m_j|}.
\]
The module retains edges where $\mathrm{IoU}_{mask}>0.10$. The algorithm extracts connected components to form reconstructed object instances. For overlapping pixels belonging to different instances, GMA applies a Winner-Takes-All procedure based on confidence scores to output a unified segmentation mask $\mathcal{M}_{sam}$.

\textbf{Geometry-Aware Fusion:} 
Finally, the software aligns $\mathcal{M}_{fg}$ to the spatial resolution of $\mathcal{M}_{sam}$. It applies a pixel-wise logical AND operation between the two masks. This step acts as a geometric filter: it removes predictions located outside the foreground vegetation region without altering the shape of the objects detected by the zero-shot segmentation branch.

\subsection{Software functionalities and interfaces}
\label{sec:functionalities}

FrontVeg V2 provides two complementary interfaces:
\begin{itemize}
    \item \textbf{Command-Line Interface (CLI):} Designed for automated batch processing on HPC clusters. Users define input/output directories, model paths, and parameters (tile size, overlap, prompt) via arguments, allowing large-scale phenotyping without code modification.
    \item \textbf{Napari Plugin:} A graphical interface (\url{https://napari-hub.org/plugins/frontvegv2.html}) for interactive analysis. Agronomists can load images, tune pipeline parameters visually, and export both segmentation masks and quantitative instance statistics (object counts, areas).
\end{itemize}





\subsection{Implementation details and code snippets}
\label{sec:implementation}

The software is implemented in Python 3.10 and relies on CUDA 12.1 for GPU-accelerated inference. The following code snippets highlight key implementation mechanisms of the pipeline.

Listing~\ref{lst:batch_pipeline} illustrates the main execution workflow. It shows how the modules are initialized externally and orchestrated to process a directory of images.

\begin{lstlisting}[style=pythonstyle, caption={Batch pipeline execution.}, label={lst:batch_pipeline}]
frontveg_pipe = FrontVegPipeline(config)
sam3_model = SAM3Predictor(checkpoint_path=args.sam3_ckpt)
sam3_tiler = SAM3TiledInference(sam3_model, tile_size=args.tile_size)
post_proc = PostProcessor()

# Processes a folder, applies depth estimation, foreground masks
frontveg_pipe.process_folder(str(subdir), output_root / "temp_frontveg")

# For each image, fuse the SAM3 segmentation with the foreground mask
id_map_raw, _ = sam3_tiler.process(pil_img, prompt=args.prompt)
final_id_map, final_mask = post_proc.get_final_colored_map(id_map_raw, mask_fv)
\end{lstlisting}

Listing~\ref{lst:hybrid_mask_assembly} details the Graph-Based Mask Assembly. To optimize computation time, the module uses a two-stage approach: a fast vectorized bounding box IoU filtering, followed by a precise mask intersection validation for remaining candidates.

\begin{lstlisting}[style=pythonstyle, caption={Hybrid mask assembly with two-stage IoU validation.}, label={lst:hybrid_mask_assembly}]
# Stage 1: Fast bounding box IoU
boxes = torch.tensor([f['bbox'] for f in fragments], dtype=torch.float32)
box_iou_matrix = box_iou(boxes, boxes).numpy()

# Stage 2: Precise mask intersection validation
for i, j in np.argwhere(box_iou_matrix > 0.05):
    if intersection > 0: # ... calculates region intersection
        mask_iou = intersection / (area1 + area2 - intersection)
        if mask_iou > 0.10:
            adjacency[i, j] = adjacency[j, i] = True

# Resolve graph
n_leaves, labels = connected_components(
csgraph=csr_matrix(adjacency), directed=False
)
\end{lstlisting}

Listing~\ref{lst:final_mask_fusion} demonstrates the final geometry-aware fusion. The logical intersection ensures that only objects physically located within the foreground vegetation row are retained in the final instance map.

\begin{lstlisting}[style=pythonstyle, caption={Final mask fusion using strict mask intersection.}, label={lst:final_mask_fusion}]
# 1. SAM Binary Mask
mask_sam = (id_map_sam > 0).astype(np.uint8) * 255

# 2. Final Binary Intersection (The "True Foreground")
final_binary_mask = cv2.bitwise_and(mask_sam, (mask_fv > 127).astype(np.uint8) * 255)

# 3. Clean ID Map: Keep the ID only if it's inside our strict intersection
final_id_map = np.where(final_binary_mask > 0, id_map_sam, 0).astype(np.uint16)
\end{lstlisting}

\section{Illustrative examples}

\begin{figure}[!ht]
    \centering
    \includegraphics[width=0.9\textwidth]{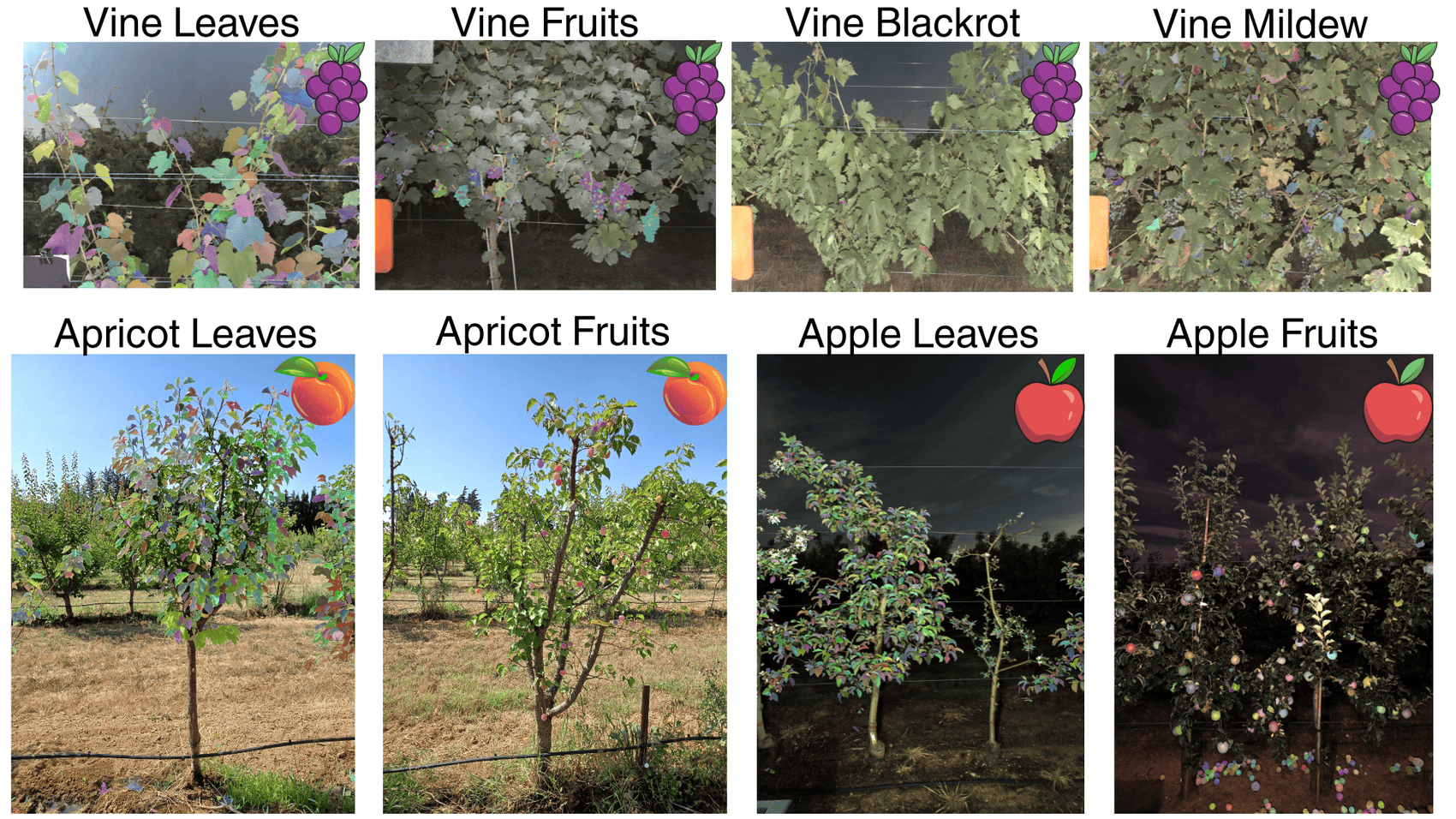}
    \caption{Zero-shot segmentation of leaves, fruits, and disease 
    symptoms (black rot and downy mildew) using trait-specific prompts.}
    \label{fig:illustrative_expl2}
\end{figure}

\begin{figure}[!ht]
    \centering
    \includegraphics[width=0.9\textwidth]{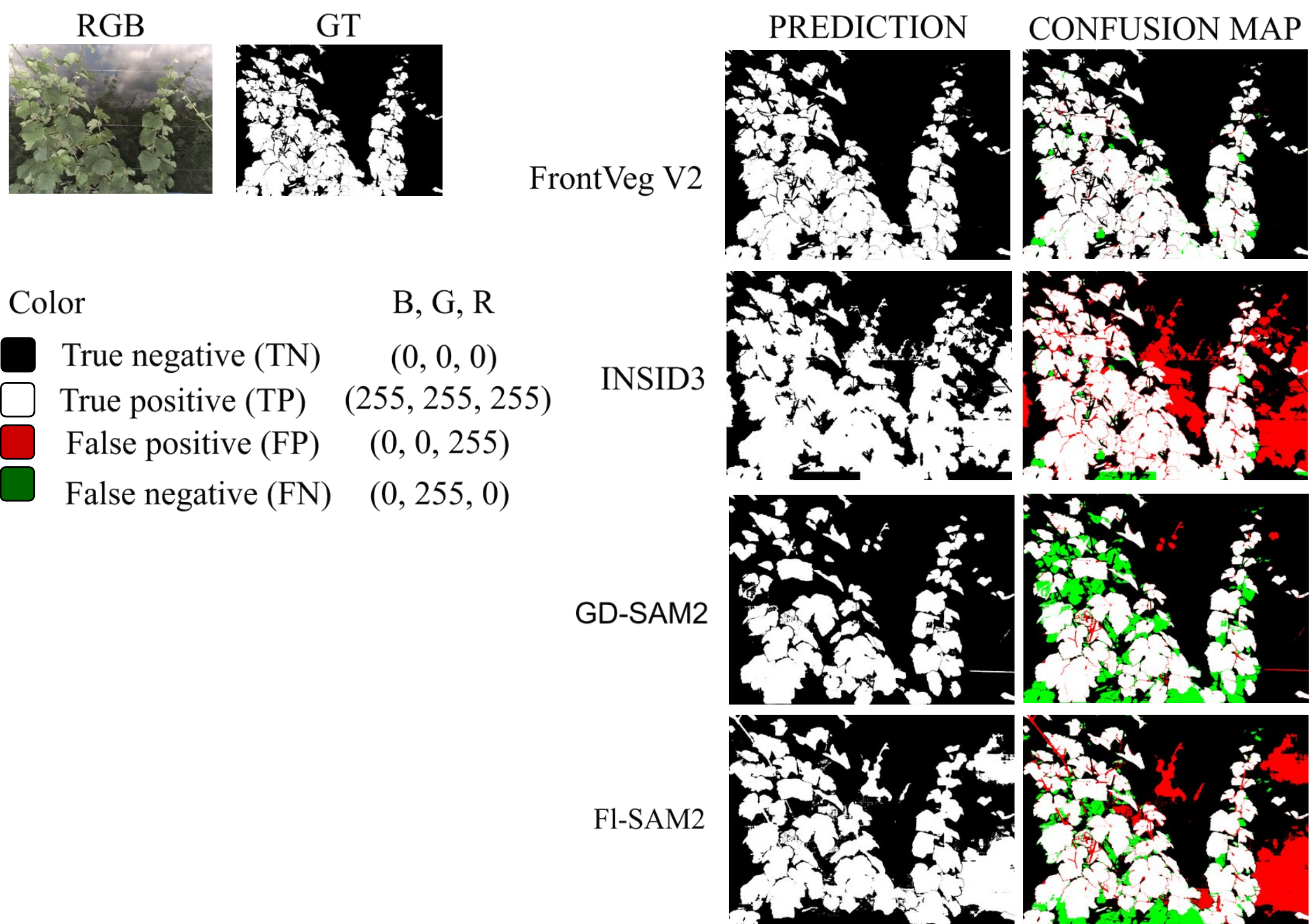}
    \caption{Qualitative comparison of leaf segmentation between FrontVeg V2 
    and existing state-of-the-art methods.}
    \label{fig:illustrative_expl}
\end{figure}

In addition to the examples shown in Fig.~\ref{fig:illustrative_expl} and Fig.~\ref{fig:illustrative_expl2}, we provide a video demonstration of the FrontVeg V2 graphical interface at: \url{https://youtu.be/0sHq0k9mMcc}.

Figure~\ref{fig:illustrative_expl2} illustrates the 
FrontVeg V2 applied to high-resolution images of trellised crops (vine, apricot, and apple) to segment foreground-aware trait including, leaves, fruits, and vine disease symptoms (blackrot and mildew).
Figure~\ref{fig:illustrative_expl} demonstrates the qualitative results of zero-shot leaves
segmentation with a comparison with existing methods.

\section{Impact}

FrontVeg V2 provides a single, training-free workflow for multi-crop plant trait segmentation, eliminating the need to annotate datasets and train dedicated models for each specific task. The software currently targets leaves, fruits, and disease symptoms (black rot, downy mildew) across grapevine, apple, and apricot orchards.

The performance impact of FrontVeg V2 was evaluated on a vineyard benchmark. Table~\ref{tab:sota_impact} summarizes the comparison with existing zero-shot pipelines \cite{hamza2026foregroundaware}. FrontVeg V2 outperforms state-of-the-art tools across all evaluated traits, with substantial gains on complex targets such as fruits (Dice 0.92 vs. 0.58 for INSID3 \cite{cuttano2026insid3}) and disease symptoms, where competing methods fail (Dice $< 0.20$).

\begin{table}[htbp]
\centering
\caption{Comparison with existing zero-shot segmentation approaches on the vineyard benchmark (mean $\pm$ standard deviation for Dice and IoU). Best results are in bold.}
\label{tab:sota_impact}
\resizebox{\textwidth}{!}{%
\begin{tabular}{lcccccccc}
\hline
& \multicolumn{2}{c}{GD-SAM2 \cite{ren2024grounding,ravi2024sam2segmentimages}} 
& \multicolumn{2}{c}{Fl-SAM2 \cite{xiao2024florence,ravi2024sam2segmentimages}}
& \multicolumn{2}{c}{INSID3 \cite{cuttano2026insid3}}
& \multicolumn{2}{c}{\textbf{FrontVeg V2}} \\
Trait & Dice & IoU & Dice & IoU & Dice & IoU & Dice & IoU \\
\hline
Leaves
& 0.84 $\pm$ 0.04 & 0.73 $\pm$ 0.06
& 0.81 $\pm$ 0.05 & 0.68 $\pm$ 0.06
& 0.90 $\pm$ 0.04 & 0.82 $\pm$ 0.06
& \textbf{0.95 $\pm$ 0.02} & \textbf{0.90 $\pm$ 0.04} \\

Fruits
& 0.41 $\pm$ 0.02 & 0.28 $\pm$ 0.17
& 0.13 $\pm$ 0.08 & 0.07 $\pm$ 0.05
& 0.58 $\pm$ 0.20 & 0.44 $\pm$ 0.20
& \textbf{0.92 $\pm$ 0.06} & \textbf{0.85 $\pm$ 0.10} \\

Black rot
& 0.00 $\pm$ 0.00 & 0.00 $\pm$ 0.00
& 0.00 $\pm$ 0.00 & 0.00 $\pm$ 0.00
& 0.00 $\pm$ 0.00 & 0.00 $\pm$ 0.00
& \textbf{0.61 $\pm$ 0.11} & \textbf{0.45 $\pm$ 0.12} \\

Downy mildew
& 0.14 $\pm$ 0.09 & 0.08 $\pm$ 0.05
& 0.12 $\pm$ 0.07 & 0.06 $\pm$ 0.04
& 0.16 $\pm$ 0.10 & 0.09 $\pm$ 0.06
& \textbf{0.73 $\pm$ 0.14} & \textbf{0.59 $\pm$ 0.16} \\
\hline
\end{tabular}%
}
\end{table}

To validate the algorithmic choices of FrontVeg V2, an ablation study was conducted (Table~\ref{tab:ablation_impact}). The results confirm that using the foundation model (SAM3) alone yields poor results on small or complex targets (e.g., Dice 0.00 on black rot). The introduction of tiled inference and mask assembly significantly improves instance retrieval, while the geometry-aware foreground filter removes background noise. Their combined integration in FrontVeg V2 yields the highest overall accuracy.

\begin{table}[htbp]
\centering
\caption{Ablation study of FrontVeg V2 software components on the vineyard benchmark (mean $\pm$ standard deviation for Dice and IoU).}
\label{tab:ablation_impact}
\resizebox{\textwidth}{!}{%
\begin{tabular}{lcccccccc}
\hline
& \multicolumn{2}{c}{Leaves}
& \multicolumn{2}{c}{Fruits}
& \multicolumn{2}{c}{Black rot}
& \multicolumn{2}{c}{Downy mildew} \\
Configuration & Dice & IoU & Dice & IoU & Dice & IoU & Dice & IoU \\
\hline
SAM3
& 0.78 $\pm$ 0.11 & 0.65 $\pm$ 0.12
& 0.78 $\pm$ 0.20 & 0.69 $\pm$ 0.20
& 0.00 $\pm$ 0.00 & 0.00 $\pm$ 0.00
& 0.35 $\pm$ 0.21 & 0.23 $\pm$ 0.15 \\

+ Tiling + NMS
& 0.90 $\pm$ 0.04 & 0.83 $\pm$ 0.07
& 0.90 $\pm$ 0.08 & 0.83 $\pm$ 0.14
& 0.50 $\pm$ 0.16 & 0.35 $\pm$ 0.15
& 0.72 $\pm$ 0.15 & 0.59 $\pm$ 0.17 \\

+ Our Foreground
& 0.78 $\pm$ 0.11 & 0.65 $\pm$ 0.13
& 0.80 $\pm$ 0.25 & 0.71 $\pm$ 0.23
& 0.00 $\pm$ 0.00 & 0.00 $\pm$ 0.00
& 0.35 $\pm$ 0.21 & 0.23 $\pm$ 0.15 \\

+ Tiling + NMS + Our Foreground
& 0.94 $\pm$ 0.03 & 0.89 $\pm$ 0.05
& \textbf{0.92 $\pm$ 0.08} & \textbf{0.86 $\pm$ 0.13}
& 0.54 $\pm$ 0.14 & 0.38 $\pm$ 0.14
& \textbf{0.73 $\pm$ 0.15} & \textbf{0.60 $\pm$ 0.17} \\

\textbf{FrontVeg V2}
& \textbf{0.95 $\pm$ 0.02} & \textbf{0.90 $\pm$ 0.04}
& \textbf{0.92 $\pm$ 0.06} & \textbf{0.85 $\pm$ 0.10}
& \textbf{0.61 $\pm$ 0.11} & \textbf{0.45 $\pm$ 0.12}
& \textbf{0.73 $\pm$ 0.15} & \textbf{0.60 $\pm$ 0.16} \\
\hline
\end{tabular}%
}
\end{table}

Beyond raw performance, FrontVeg V2 resolves the manual annotation bottleneck in digital phenotyping. Creating the 158-image evaluation dataset required approximately 200~hours of manual expert annotation. By offering zero-shot inference without fine-tuning, FrontVeg V2 saves an estimated 1,300~hours of manual labor per 1,000 images, accelerating phenotypic data collection.

Earlier versions of FrontVeg have supported agricultural research, including doctoral studies on grapevine leaf-area estimation and published work on tree segmentation \cite{metuarea2025individualseg}. The initial version, released in April, reached 5,789 downloads on PyPI during its first year, while FrontVeg V2 registered 2,666 downloads within its first month (August 2026) on PyPI representing a five-fold increase in average daily downloads.

\section{Conclusions}

FrontVeg V2 provides a training-free framework for foreground-aware plant trait segmentation in high-resolution images of trellised crops. By combining monocular depth estimation with tiled zero-shot segmentation and graph-based mask assembly, the software reduces interference from neighboring vegetation rows while preserving fine plant structures. Its modular implementation and availability through both command-line and Napari interfaces make it suitable for interactive analysis and large-scale phenotyping workflows. FrontVeg V2 therefore offers a reusable basis for geometry-aware, multi-crop and multi-trait phenotyping without task-specific model retraining. The pipeline runs on consumer-grade hardware (Intel i7-11850H, 32 GB RAM, RTX 3080 16 GB) without cloud dependency. VADT preprocessing takes ~1 s/image, while sliding-window SAM3 inference remains the main bottleneck (~2 min/class), providing a baseline for future real-time edge deployment.


\section*{Acknowledgements}
\label{}

This work was supported by The French Programme et Équipements Prioritaires de Recherche (PEPR) AgroEcoNum (\url{https://www.pepr-agroeconum.fr/}), the France 2030 program through the Agence Nationale de la Recherche (ANR), and the European Union’s Horizon Europe research innovation programme (No. 101094587, PHENET). This research used resources of the GLiCID Computing Facility (Ligerien Group for Intensive Distributed Computing, \url{https://doi.org/10.60487/glicid} , Pays de la Loire, France). This project was provided with computing HPC and storage resources by GENCI at IDRIS thanks to the grant 2025-AD010115553R1 on the supercomputer Jean Zay’s H100 partition. We would like to thank Sirine Gharbi, Oumaima Karia, Justin Langlois,  Paul Persello, and Thomas Lebouc, for their valuable assistance in annotating the dataset used in this study.






\end{document}